# Automatic Three-Dimensional Cephalometric Annotation System Using Three-Dimensional Convolutional Neural Networks

Sung Ho Kang, MS[1], Kiwan Jeon, PhD[2], Hak-Jin Kim, DDS, PhD[3], Jin Keun Seo, PhD[4],

and Sang-Hwy Lee, DDS, PhD[5]

[1] National Institute of Mathematical Science, Division of Integrated Mathematics; KT Daeduk 2 Research Center, 70 Yuseong-daero, 1689 beon-gil, Yuseong-Gu, Daejeon, (305-811) Republic of Korea. runits@nims.re.kr
[2] National Institute of Mathematical Science, Division of Integrated Mathematics; KT Daeduk 2 Research Center, 70 Yuseong-daero, 1689 beon-gil, Yuseong-Gu, Daejeon, (305-811) Republic of Korea. jeonkiwan@nims.re.kr
[3] Department of Oral and Maxillofacial Surgery, Oral Science Research Center, College of Dentistry, Yonsei University; 225 Kumhak-No, Cheonin-Gu, Gyeonggi-Do (17046), Republic of Korea. OMSKIM@yuhs.ac
[4] Department of Computational Science and Engineering, Yonsei University; 50, Yonsei-ro Seodaemun-gu, Seoul, (03722) Republic of Korea. seoj@yonsei.ac.kr
[5] Dept of Oral and Maxillofacial Surgery and Oral Science Research Center, College of Dentistry, Yonsei University; 50-1 Yonsei-ro, Seodaemun-gu, Seoul (03722) Republic of Korea. sanghwy@yuhs.ac; sanghwy@gmail.com

**Keywords: Cephalometry, Annotation, Deep Learning, Convolutional Neural Network, Three-dimensional, Computed tomography**

## Abstract

Background: Three-dimensional (3D) cephalometric analysis using computerized tomography data has been rapidly adopted for dysmorphosis and anthropometry. Several different approaches to automatic 3D annotation have been proposed to overcome the limitations of traditional cephalometry. The purpose of this study was to evaluate the accuracy of our newly-developed system using a deep learning algorithm for automatic 3D cephalometric annotation.

Methods: To overcome current technical limitations, some measures were developed to directly annotate 3D human skull data. Our deep learning-based model system mainly consisted of a 3D convolutional neural network and image data resampling.

Results: The discrepancies between the referenced and predicted coordinate values in three axes and in 3D distance were calculated to evaluate system accuracy. Our new model system yielded prediction errors of 3.26, 3.18, and 4.81 mm (for three axes) and 7.61 mm (for 3D). Moreover, there was no difference among the landmarks of the three groups, including the midsagittal plane, horizontal plane, and mandible ($p>0.05$).

Conclusion: A new 3D convolutional neural network-based automatic annotation system for 3D cephalometry was developed. The strategies used to implement the system were detailed and measurement results were evaluated for accuracy. Further development of this system is planned for full clinical application of automatic 3D cephalometric annotation.

# Background

Cephalometry has played a central role in craniofacial measurement and analysis since its introduction in 1930s(Broadbent, 1931; Hofrath, 1931) being applied in diverse medical and biological fields, including craniofacial dysmorphosis, clinical dentistry, growth, anthropology, and comparative anatomy.(Byrum, 1971; Kolotkov, 1973; Reitzik, 1972) Its two-dimensional (2D) radiographic image has several disadvantages such as the difficulty in identifying the superimposed anatomical structures. Three-dimensional (3D) cephalometric analysis with computed tomographic images is gaining in popularity because it addresses these limitations.

Cephalometric analysis is achieved through cephalometric annotation, i.e., landmark detection of significant anatomical structures. This process demands a high level of experience and expertise as well as considerable time. These difficulties are reinforced as 2D analysis transits to 3D, drastically increasing data volume and geometric complexities. 3D cephalometry basically demands more working time and a greater learning curve than does 2D cephalometry, posing a major obstacle to 3D cephalometry.

Several approaches have been suggested to overcome the limitations of 3D cephalometry, one effective solution being automatic 3D annotation. Implementations have drawn on various algorithms, including knowledge-based, model-based and machine learning-based approaches.(Codari, Caffini, Tartaglia, Sforza, & Baselli, 2017; Gupta, Kharbanda, Sardana, Balachandran, & Sardana, 2015; Makram & Kamel, 2014; Montufar, Romero, & Scougall-Vilchis, 2018; Shahidi, Oshagh, Gozin, Salehi, & Danaei, 2013) The knowledge- or model-based approaches were initially adopted due to their ease of development. However, their high ambiguity in identifying complex craniofacial structure hindered wider application. More recent developments in machine learning-based, particularly the deep learning-based approach, hold promise for addressing ambiguity.(Litjens et al., 2017) This approach may be more useful in the context of 3D medical imaging due to its higher efficiency.(Kang, Min, & Ye, 2017) We thus here propose a convolutional neural network (CNN) deep learning algorithm as a solution to the challenge of automated full 3D cephalometric annotation. Our search of the literature reveals no trials of such machine learning-based automatic 3D cephalometric analysis.

The main issues in developing machine learning-based 3D cephalometry with 3D volume data are the increased number of parameters, the need for high performance computing, and greater computational complexity.(de Brébisson & Montana, 2015; Li et al., 2014; Lyksborg, Puonti, Agn, & Larsen, 2015; Prasoon et al., 2013; Roth et al., 2014) We aimed to overcome these obstacles by introducing a resampling method which would help improve the computational efficiency while maintaining the global view of 3D structure.

To perform the automated annotation of cephalometric landmarks on 3D computed tomographic image data, we created a model system that employs resampling in conjunction with the CNN algorithm. The purpose of this study was to develop and to evaluate the accuracy of our new model system, which uses a deep learning algorithm to locate 3D cephalometric landmarks on the craniofacial skeletal structure.

## Materials and Method

### Subjects

The anonymized CT data from our previous 3D cephalometric study of normal subjects(Lee et al., 2014) was used. 27 normal Korean adults with skeletal class I occlusion (17 males and 10 females; 24.3 years old for males and 23.5 years old for females) volunteered for the study. Both clinical and cephalometric examinations with a dental cast were used to rule out facial dysmorphosis or malocclusion. Subjects with any facial disharmony, facial asymmetry, or history of surgical or orthodontic treatment were excluded. The original work was approved by the local Ethics Committee of the Dental College Hospital, Yonsei University, Seoul, Korea (IRB number: 2-2009-0026), and informed consent was obtained from each subject. The subjects were divided into two groups including the learning data group (n=18) and the test data group (n=9).

### Image acquisition

Each subject underwent CT imaging with the Frankfort Horizontal (FH) line perpendicular to the floor using a high-speed Advantage CT system (GE Medical System, Milwaukee, USA) from the soft tissue chin to vertex of skull using a high-resolution bone algorithm protocol (200mA, 120kV, scanning time 1s, a 512×512pixel reconstruction matrix). The CT image data were saved in the DICOM file format, transferred to a personal computer and reconstructed to 3D skull images using SimPlant Pro (Materialise Dental, Leuven, Belgium).

### Landmark localization and coordinate conversion

Two of the authors, each having done 3D cephalometry for more than ten years in a university hospital setting, independently used SimPlant software to locate 12 landmark points for 3D cephalometric analysis. Landmark point detail and definitions are provided in Supplementary Table 1. The coordinate value of each landmark point was converted from a SimPlant coordinate to a DICOM voxel coordinate to construct the label data for deep learning.

The cephalometric landmarks for this study included nasion, bregma, foramen magnum (center), menton, mandibular foramen, coronoid, orbitale and porion, the latter four points being bilateral (Supplementary Table 1).

The former three points were used to determine the midsagittal plane, the latter two bilateral points to determine the Frankfort plane. The landmarks were then divided into three groups: midsagittal plane, horizontal plane, and the mandible. The x-axis indicated the transverse dimension, the y-axis the anterior-posterior dimension and the z-axis the superior-inferior dimension.

**Accuracy measurement for landmark localization**

The distance between the referenced and predicted coordinate value was measured for each landmark point in the x-, y-, and z-axis as well as in 3D distance. Regarding the referenced coordinate value, two of the authors performed the landmark pointing two times at a one-week interval for each subject. Their discrepancy was calculated to have an intraclass correlation coefficient (ICC) between and within observers with 95% confidence intervals.

# Results

**Construction of training set and pre-processing**

One of the main drawbacks in utilizing medical image data is the limited number of learning data sets, since labelling is time-consuming and stringent regulations hinder the free utilization of medical data. The limited data sets for learning and the labelling of single pixels inevitably decrease the sensitivity due to the higher bias, lower variance, and bias-variance tradeoff.(de Vos et al., 2017) We therefore designed a new learning system using a training set and pre-processing to avoid these shortcomings; the first strategy involved data labelling method: a mathematical function was used to characterize landmark locations in terms of a smooth decay to neighboring coordinates (Fig 1A). This was done to prevent overfitting and to provide general learning performance despite the small training data set. Figure 1A shows the 3D coordinates of an example landmark point located at (64, 82, 24) for (x, y, z). Our second strategy addressed the lack of computing resources by downsizing the image data through resampling to reduce computational demands while increasing learning predictability. After trials of several different sizes, we resized 3D input data to obtain a new 2 mm pixel at each axis. Figure 1B shows the sagittal views of the skull in the original (1B-a) and resampled images (1B-b) from the same CT data. The new pixel spacing following data resampling incurred size differences on the x-, y- and z-coordinate axes, as shown in figure 1C. We resolved this issue by padding the pixels to a size that could reset the data size of 128x128x152. The normalization was performed between Hounsfield values of -1000 and 400.

Our final strategy was to augment the 3D data by the application of the same parameters. After test applications using several different values, augmentation parameters within 15% were applied for the translation and rotation in 3D axis for this study.

**CNN architecture and its training**

The model system was designed to archive the characteristics of 3D input data set using 3D convolutional layer and to classify them to the local coordinates of the landmark point using the fully-connected layers. Figure 2 shows the proposed architecture of the model system; 3D input data size was 128x128x158, and the features were extracted by the four-layered 3D convolution. 3D convolution was basically performed with 3x3x3 kernels, and the max-out activation (k=2) was applied for non-linear activation. 3D convolution adopted the serial reduction of axis dimension as (1,3,3)(3,1,3)(3,3,1) for (x, y. z) axis. After 3D convolution, max-pooling was performed; the data after the four convolutional layers and fully-connected layer were assigned to each coordinate value of the landmark. The softmax activation was applied to the end of fully-connected layer for each landmark points. The dropout layer was adopted to prevent the overfitting in the block connection, and the optimizer was chosen by adadelta (Zeiler, 2012).

**Landmark localization accuracy**

The predicted coordinate values for all landmarks were calculated by the developed algorithms and distances between referenced and predicted coordinate values were measured in the x-, y-, and z-axis as well as in 3D distance. Table 1 shows the mean distance (with standard deviation) between the values. The landmark point foramen magnum (center) showed the highest level of accuracy with 6.37 mm of error in 3D distance, the point right orbitale being the worst with 9.57 mm in 3D. All the points ranged within relatively small error boundaries of 1.79 mm and 7.15 mm (1.79 mm for y-axis and 7.15 mm for z-axis of right orbitale) in 2D. The average 3D distance was 7.81 mm for midsagittal plane landmarks, 7.41 mm for the horizontal plane points, and 7.66 mm in the mandible. There were no significant differences between the three groups (by Kruskal-Wallis test; $p>0.05$; details not shown). The total mean 3D difference for all landmarks was 7.61 mm. In addition, the 2D discrepancies at the three axis coordinates were well within a range of 2-5 mm, the average 2D difference for the x-, y-, and z-axis being 3.26, 3.18, and 4.81 mm (with 3.75 mm of total 2D difference). A statistical analysis of the measured and referenced coordinate values on the x-, y-, and z-axes was made to show that all except the y-axis value of the left orbitale were not significantly different (by Wilcoxon signed-rank test; $p>0.05$ for all variables except left orbitale; details not shown).

The distribution of the discrepancy, shown in the box plots in Figure 3, indicate that most landmark points fell within a certain range. The point porion seemed to be an easily predicted landmark, while the point bregma and nasion were difficult to predict. Figure 4 shows the positions of the sampled landmark points on 3D skull models. The figure shows that some predicted landmark points, such as nasion, closely match the referenced landmarks, while others differ somewhat from the referenced ones.

**Methods of error for landmark localization**

The referenced coordinate values were produced from the average value of measurement results made by two observers. The calculated mean discrepancy for all landmark points was 0.49, 1.02, 1.40, and 1.80 mm in x-axis, y-axis, z-axis, and 3D. ICC for intra-observer was 0.95 (as Cronbach's alpha) and for inter-observer was 0.92.

## Discussion

The purpose of this study was to develop a new model system utilizing a deep learning algorithm to automatically locate cephalometric landmarks on the craniofacial skeletal structure and to evaluate its accuracy. Specifically, we created a deep learning-based 3D automatic annotation system, mainly using resampling and 3D CNN, for 3D craniofacial CT data.

Many studies have reported on automatic annotation systems for 2D cephalometry.(Forsyth & Davis, 1996; Kaur & Singh, 2015; Leonardi, Giordano, Maiorana, & Spampinato, 2008; Shahidi et al., 2013; Tanikawa, Yagi, & Takada, 2009) These employed various artificial intelligence or machine learning algorithms, with recent research including knowledge-based, model-based, soft computing-based, and hybrid approaches.(Cardillo & Sid-Ahmed, 1994; Giordano, Leonardi, Maiorana, Cristaldi, & Distefano, 2005; Hutton, Cunningham, & Hammond, 2000; Levy-Mandel, Venetsanopoulos, & Tsotsos, 1986; Parthasarathy, Nugent, Gregson, & Fay, 1989; Rudolph, Sinclair, & Coggins, 1998; Vučinić, Trpovski, & Šćepan, 2010) Early studies utilized knowledge-based approaches using edge detection and image-processing techniques.(Giordano et al., 2005; Levy-Mandel et al., 1986; Parthasarathy et al., 1989; Vučinić et al., 2010) Edge detection employing contour extraction and landmark localization on the edge might fail to register morphological variability, depended on the quality of the input images, and is sensitive to image noise. However, it was faster and easier to perform on landmarks with extracted edges.

Model-based approaches in 2D cephalometry use a reference model (template) with landmarks.(Cardillo & Sid-Ahmed, 1994; Hutton et al., 2000; Rudolph et al., 1998) They have been the most popular method due to their accommodation of shape variability and usefulness in establishing landmarks on definite structures. However, they require reference models obtained by averaging morphologically variable craniofacial structures and also by performing constrained deformation, both of which could result in inaccurate modeling.

Soft computing-based approaches included fuzzy logic and machine or deep learning.(Arik, Ibragimov, & Xing, 2017; Chakrabartty, Yagi, Shibata, & Cauwenberghs, 2003; Douglas, 2004; Innes, Ciesielski, Mamutil, & John, 2002; Leonardi et al., 2008; Lindner et al., 2016; Sanei, Sanaei, & Zahabsaniei, 1999) Earlier studies focused on the fuzzy logic algorithm(Douglas, 2004; El-Feghi, Sid-Ahmed, & Ahmadi, 2004) whereas recent studies have relied mainly on neural networks and deep learning.(Arik et al., 2017; Chakrabartty et al., 2003; Giordano et al., 2005; Innes et al., 2002; Lindner et al., 2016; Pei, Liu, Zha, Han, & Xu, 2013) These could accommodate morphological variation and were tolerant of noise, but their accuracy levels depended on the training set and the network parameters needed to be determined by experience.

Deep learning is a part of machine-learning, which mimics information processing and communication patterns in a biological nervous system. It has shown outstanding performance in solving many problems in computer vision and biomedical applications.(Giordano et al., 2005; LeCun, Bengio, & Hinton, 2015; Wang et al., 2016) Lindner et al.(Lindner et al., 2016) reported an automatic 2D cephalometric annotation system using deep learning which demonstrated excellent performance in the 2015 Grand Challenge.(Wang et al., 2016) They used the random forest regression-voting algorithm for skull position, scale, and orientation, and the constrained local model framework for localization of landmarks, achieving an average point-to-point error of 1.66 mm with improved performance in terms of increased patch size, increased training sampling range and decreased search range.

Arik et al.(Arik et al., 2017), conducted deep learning-based 2D cephalometry using CNN, a rapidly developing deep learning algorithm that uses a variation of multilayer perceptrons(LeCun, 2015), inspired by the connectivity of the biological nervous system.(Huang & LeCun, 2006) It is especially suitable for image processing and recognition in that it exploits spatial correlations by imposing local connectivity patterns. Our search of the literature has not found any CNN-based 3D cephalometric annotation system.

Most previous automatic 3D cephalometric annotation studies employed model-based approaches which depended on a feature-extracted reference model system.(Codari et al., 2017; Gupta et al., 2015; Makram & Kamel, 2014; Montufar et al., 2018) The geometrical shape and structure of each data set is unique due to the craniofacial structural variations, making it difficult to detect the precise location of landmarks. The CNN algorithm was selected mainly because it employs a hierarchical structure to propagate information on salient features to subsequent layers while exploiting spatially local correlations. Being restricted to specified raw images, other algorithms could not be well-generalized for challenging the image recognition tasks with complex craniofacial patterns.

With regard to automatic annotation systems for 2D cephalometry, it has been advocated that the positional difference between the referenced and predicted (measured) landmarks should be less than 2 mm in a clinical environment, though less than 4 mm would be acceptable(Shahidi et al., 2013). We know that this recommendation cannot be directly applied to 3D cephalometry and indeed, there has been no work to standardize automatic 3D cephalometric annotation. Further study is necessary to calibrate a 3D annotation system.

The error ranges reported by previous 3D cephalometric annotation studies all seem excellent; Gupta et al.(Gupta et al., 2015; Gupta, Kharbanda, Sardana, Balachandran, & Sardana, 2016) searched for 20 cephalometric landmarks in conebeam CT images using knowledge-based approaches. They achieved an average accuracy of 2.01 mm, 64.67% of the landmarks falling within a range of 0 to 2 mm and the highest mean error in the linear measurements being 2.63 mm. Shahidi et al.(Shahidi et al., 2013) evaluated 14 cephalometric landmarks in conebeam CT by model-based rigid registration and reported that 63.57% of landmarks had an average error less than 3 mm. Makram and Kamel(Makram & Kamel, 2014) proposed a localization system of 20 landmarks using model-based approaches with rigid and elastic registration, 90 percent of their landmarks having a localization error less than 2 mm.

Codari et al.(Codari, Caffini, Tartaglia, Sforza, & Baselli, 2016; Codari et al., 2017) introduced a localization method using automatic segmentation and template-based non-rigid holistic registration which yielded an average localization error of 1.99 mm on 21 points. Finally, Montufar, J., et al. (2018)(Montufar et al., 2018) used 24 conebeam CT scans and their orthogonal coronal and sagittal projections to obtain active shape model-based registration for 18 landmarks, achieving a 3.64-mm mean error; the highest localization errors were found in the porion and sella regions.

Our results showed 7.61 mm of measurement error in average 3D distance between the predicted and referenced position value for the localization of cephalometric landmarks, and 3.26, 3.18, and 4.81 mm (for the x-, y-, and z-axis) in average 2D distance. These errors fell between 6.37 and 9.57 mm in 3D coordinate and 1.79 and 7.15 mm in 2D. These differences came to 1-5 voxels of erroneous prediction in that one voxel came to be 2 mm in one dimension. It is true that these error levels in 3D are somewhat greater than in previous studies and insufficient for clinical application(Codari et al., 2017; Gupta et al., 2015; Makram & Kamel, 2014; Montufar et al., 2018; Shahidi et al., 2013), mainly due to the limited computational resource, but also to the limited training data. Our 2D distance error results indicate less distortion than in previous 2D cephalometric annotation reports(Kaur & Singh, 2015; Leonardi et al., 2008; Shahidi et al., 2013; Tanikawa et al., 2009), and hold promise for further improving the model system for clinical settings.

In addition, the landmarks on the horizontal plane were not significantly different from those of the midsagittal plane and mandible. Moreover, there was no difference between the three groups of landmarks ($p>0.05$; details not shown), indicating a favorable insensitivity to the structural complexity of landmarks. In previous reports, horizontal plane landmarks included in our system such as the porion point posed greater difficulty for automatic annotation due to geometrical complexity.(Montufar et al., 2018)

We admit that the results of this study clearly suggest the need for further development of our model system, which lacks sufficient accuracy for clinical application. In addition, our system was not good at the accurate localization of points on the skeletal surface, values on the three axes not correlating with each other at a specific landmark. However, our current model is the first to utilize 3D CNN for automatic annotation of 3D cephalometric analysis and can serve as an approximate guide to landmarks in a region of interest to reduce the time needed for annotation.

## Conclusions

The purpose of this study was to develop an automated annotation system for 3D cephalometric landmarks based on a deep learning algorithm. To achieve this, we introduced a new 3D CNN-based automatic annotation system for 3D cephalometry. The strategies used to implement the system were detailed and measurement results were evaluated for accuracy. The results of this study clearly showed lacks sufficient accuracy for clinical application, especially for accurate localization of points on the skeletal surface. Further development of this system will lead to the full clinical application of automatic 3D cephalometric annotation.

**Table 1.** The discrepancy between the referenced and predicted positions of landmarks on the x-, y-, and z-axis as well as in 3D. (unit in mm)

| Group | Landmark | 3D Distance* | Distance* | | | Group Average Distance |
|---|---|---|---|---|---|---|
| | | | x-axis | y-axis | z-axis | |
| Midsagittal plane | CFM | 6.59 ± 2.73 | 2.28 ± 1.97 | 3.05 ± 2.56 | 4.64 ± 2.33 | 7.81 |
| | Bregma | 9.37 ± 3.49 | 4.59 ± 2.14 | 5.36 ± 4.00 | 4.92 ± 2.65 | |
| | Na | 7.47 ± 6.17 | 2.63 ± 3.03 | 1.83 ± 1.09 | 5.96 ± 6.26 | |
| Horizontal plane | R Or | 9.57 ± 3.49 | 4.08 ± 2.75 | 1.79 ± 2.65 | 7.15 ± 4.55 | 7.41 |
| | L Or | 6.66 ± 2.45 | 3.86 ± 2.74 | 3.17 ± 1.86 | 3.69 ± 2.45 | |
| | R Po | 6.37 ± 2.59 | 2.83 ± 2.22 | 3.29 ± 2.38 | 3.22 ± 2.96 | |
| | L Po | 7.05 ± 2.79 | 4.07 ± 2.26 | 1.94 ± 2.04 | 4.04 ± 3.62 | |
| Mandible | Me | 8.55 ± 4.74 | 4.03 ± 3.19 | 4.92 ± 3.40 | 4.60 ± 3.68 | 7.66 |
| | R Cor | 7.18 ± 4.77 | 3.04 ± 2.35 | 3.20 ± 4.33 | 4.30 ± 3.71 | |
| | L Cor | 7.65 ± 2.65 | 2.48 ± 1.94 | 2.32 ± 1.79 | 5.78 ± 3.92 | |
| | R F | 8.11 ± 3.79 | 3.36 ± 2.37 | 4.46 ± 2.46 | 4.63 ± 4.18 | |
| | L F | 6.79 ± 3.63 | 1.91 ± 2.13 | 2.87 ± 1.82 | 4.78 ± 4.27 | |
| Total | Average | 7.61 ± 3.61 | 3.26 ± 2.42 | 3.18 ± 2.53 | 4.81 ± 3.62 | 7.61 |

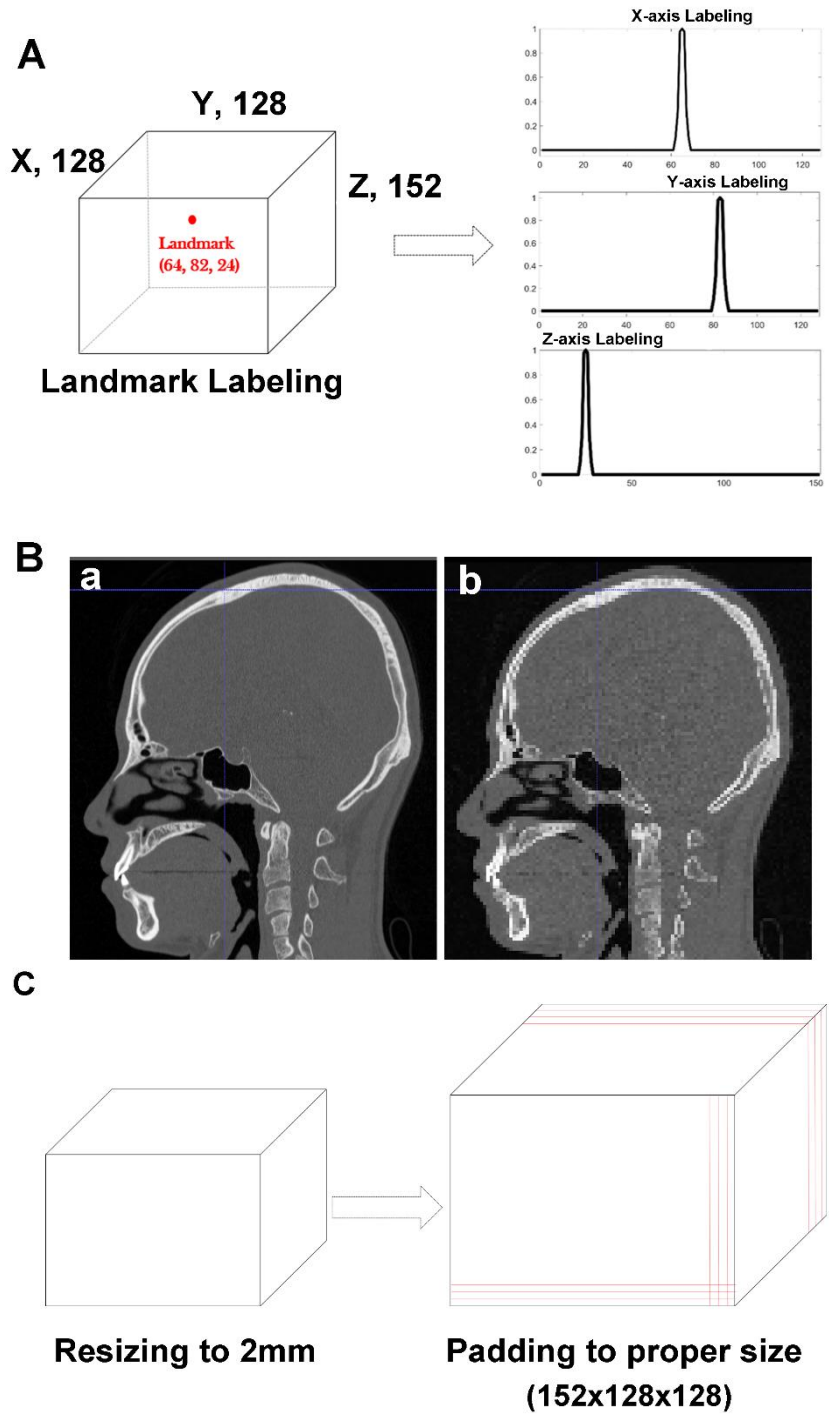

**Fig 1.** The main strategies of this study for automatic 3D cephalometric annotation. (A) A mathematized function was used to characterize sample landmark locations in terms of a smooth decay to neighboring coordinates (for an example landmark point located at (64, 82, 24) for (x, y, z)). (B) the comparison of image data in original (a) and down-sized (b) resolution. (C) the resampling-induced disproportional pixel ratio was addressed by padding the pixels to proper size.

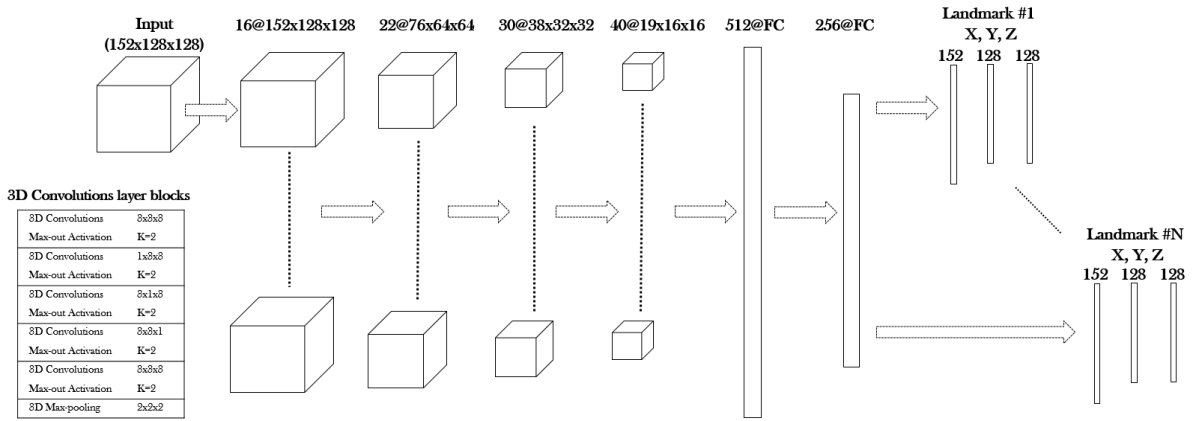

**Fig 2.** The design of CNN model system used in this study, employing 3D convolution, max poolling, and softmax activation.

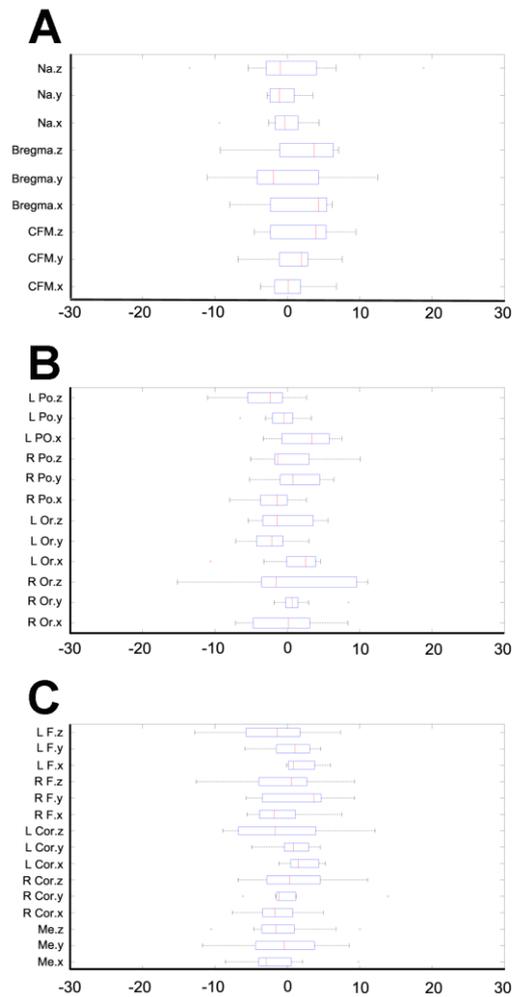

**Fig 3.** The box plot for the distribution of discrepancy between predicted and referenced coordinate values for three groups of landmarks (the midsagittal plane, horizontal plane, and mandible group).

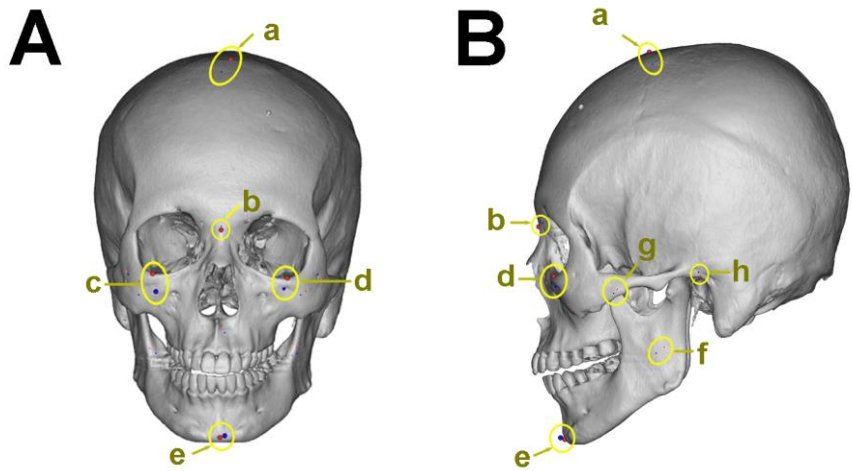

**Fig 4.** Three-dimensional skull model with the referenced and predicted landmark points to determine error levels following automatic 3D cephalometric annotation The skull model in frontal (A) and left lateral (B) view is presented with the referenced (red points) and predicted landmark points (blue), with yellow circles for each landmark point. The landmark points include included bregma (a), nasion (b), right and left orbitale (c and d), menton (e), mandibular foramen (f), coronoid (g), and porion (h).

**Supplementary Table 1.** The landmark points used in this study.

| Group | Landmark | Description | Bilaterality |
|---|---|---|---|
| Midsagittal plane | Nasion (Na) | Junction of frontal and nasal bones on the midline. | |
| | Bregma | Intersection of coronal and sagittal suture of cranial vault. | |
| | center of foramen magnum (CFM) | Midpoint of foramen magnum at the level of basion. | |
| Horizontal plane | Orbital (Or) | The most inferior point of the orbital rim. | yes |
| | Porion (Po) | The most superior point of the external auditory meatus. | yes |
| Mandible | Coronoid point (Cor) | The most distal tip point of the coronoid process. | yes |
| | Mandibular foramen (F) | The most inferior point of fossa at the opening of mandibular foramen. | yes |
| | Menton (Me) | The most inferior point of the mandibular symphysis. | |